\documentclass[10pt,twocolumn,letterpaper]{article}

\usepackage[pagenumbers]{cvpr} %

\usepackage[dvipsnames]{xcolor}

\definecolor{cvprblue}{rgb}{0.21,0.49,0.74}
\usepackage[pagebackref,breaklinks,colorlinks,citecolor=cvprblue]{hyperref}
\usepackage{bm}

\definecolor{cb-black}      {RGB}{  0,   0,   0}
\definecolor{cb-blue-green} {RGB}{  0,  073,  073}
\definecolor{cb-green-sea}  {RGB}{  0, 146, 146}
\definecolor{cb-rose}       {RGB}{255, 109, 182}
\definecolor{cb-salmon-pink}{RGB}{255, 182, 119}
\definecolor{cb-purple}     {RGB}{ 73,   0, 146}
\definecolor{cb-blue}       {RGB}{ 0, 109, 219}
\definecolor{cb-lilac}      {RGB}{182, 109, 255}
\definecolor{cb-blue-sky}   {RGB}{109, 182, 255}
\definecolor{cb-blue-light} {RGB}{182, 219, 255}
\definecolor{cb-burgundy}   {RGB}{146,   0,   0}
\definecolor{cb-brown}      {RGB}{146,  73,   0}
\definecolor{cb-clay}       {RGB}{219, 209,   0}
\definecolor{cb-green-lime} {RGB}{ 36, 255,  36}
\definecolor{cb-yellow}     {RGB}{255, 255, 109}

\newcommand{\et}[2]{${#1}^{\pm{#2}}$}
\newcommand{\etb}[2]{$\mathbf{{#1}}^{\pm{#2}}$}

\newcommand{\ets}[2]{$\underline{{#1}}^{\pm{#2}}$}

\newcommand{\PAR}[1]{\vskip4pt \noindent{\bf #1~}}

\newcommand\reallywidehat[1]{%
\savestack{\tmpbox}{\stretchto{%
  \scaleto{%
    \scalerel*[\widthof{\ensuremath{#1}}]{\kern-.6pt\bigwedge\kern-.6pt}%
    {\rule[-\textheight/2]{1ex}{\textheight}}%
  }{\textheight}%
}{0.5ex}}%
\stackon[1pt]{#1}{\tmpbox}%
}

\newcommand{\acy}[0]{HMD\textsuperscript{2}}
\newcommand{\inst}[1]{\textsuperscript{#1}}

\newcommand\blfootnote[1]{%
  \begingroup
  \renewcommand\thefootnote{}\footnote{#1}%
  \addtocounter{footnote}{-1}%
  \endgroup
}

\title{\vspace{-10mm}\acy: Environment-aware Motion Generation from Single Egocentric Head-Mounted Device \vspace{-5mm}}

\author{
Vladimir Guzov*\inst{\ddag 1,2} \and 
Yifeng Jiang*\inst{\dag 3} \and
Fangzhou Hong\inst{\ddag 4} \and 
Gerard Pons-Moll\inst{1,2} \and 
Richard Newcombe\inst{5} \and 
C. Karen Liu\inst{3} \and 
Yuting Ye\inst{5} \and 
Lingni Ma\inst{5}
}

\begin{document}

\makeatletter
\let\@oldmaketitle\@maketitle
\let\oldmaketitle\@maketitle
\renewcommand{\@maketitle}{

	\@oldmaketitle
 
 \vspace{-10mm}
	\begin{center}

\inst{1}Tübingen AI Center, University of Tübingen\qquad
\inst{2}Max Planck Institute for Informatics, Saarland Informatics Campus\qquad
\inst{3}Stanford University\qquad
\inst{4}Nanyang Technological University\qquad
\inst{5}Meta Reality Labs Research

    \vspace{3mm}

    \href{https://hmdsquared.github.io}{https://hmdsquared.github.io}

    \vspace{3mm}

 	\includegraphics[width=0.90\textwidth]{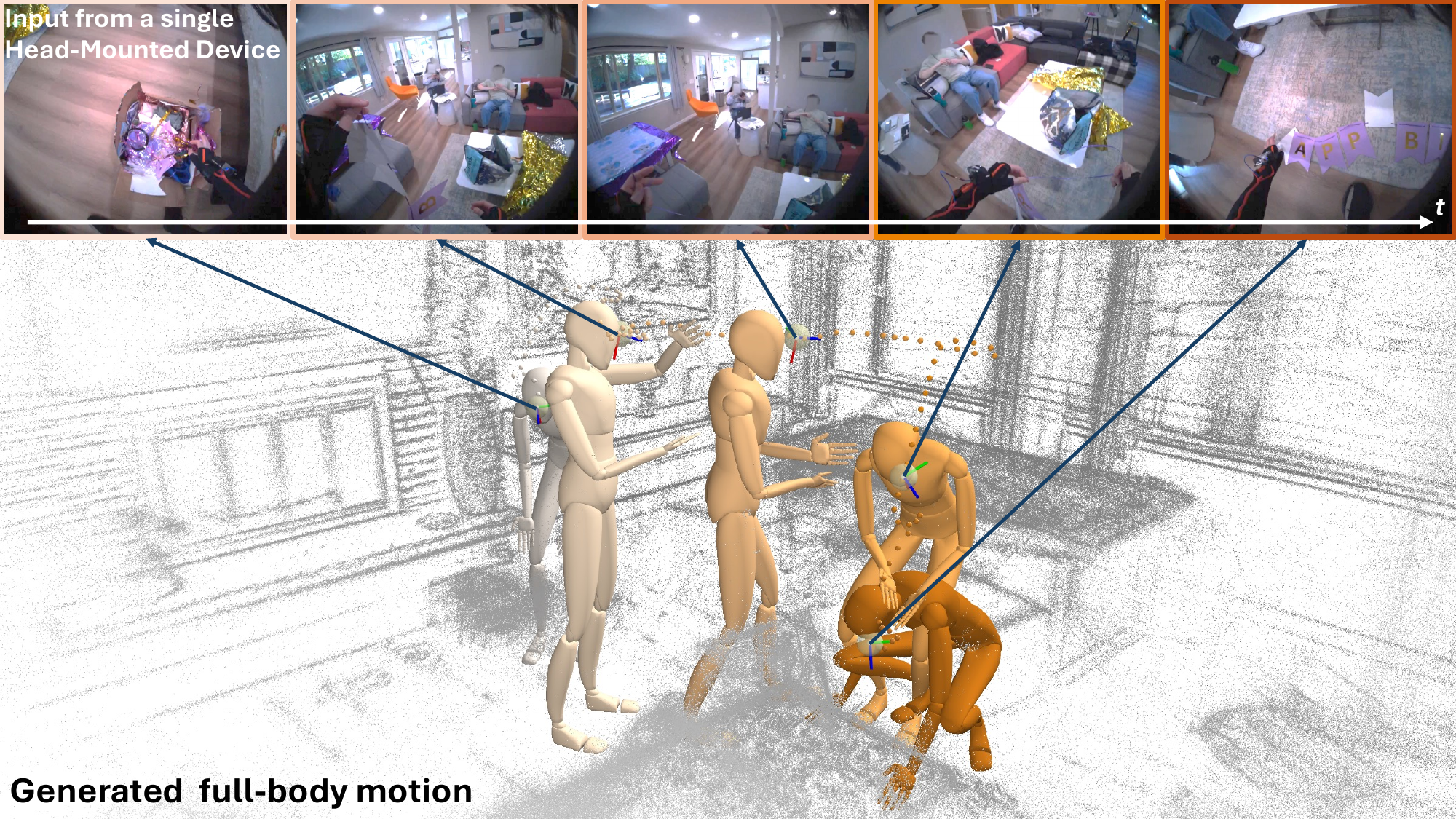}
	\end{center}
	\vspace{-3mm}
    \refstepcounter{figure}\normalfont
    Figure~\thefigure. We propose \acy, the first system for the online generation of full-body motion using a single head-mounted device (\eg Project Aria Glasses) equipped with an outward-facing camera in complex and diverse environments.
	\label{fig:teaser}
	\newline
}
\makeatother
\maketitle

\begin{abstract}
\blfootnote{* Equal contribution.}
\blfootnote{\ddag Work done during internships at Meta Reality Labs Research.}
\blfootnote{\dag Work done partially during internship at Meta Reality Labs Research.}

\vspace{-5mm}
This paper investigates the generation of realistic full-body human motion using a single head-mounted device with an outward-facing color camera and the ability to perform visual SLAM. To address the ambiguity of this setup, we present \acy, a novel system that balances motion reconstruction and generation. From a reconstruction standpoint, it aims to maximally utilize the camera streams to produce both analytical and learned features, including head motion, SLAM point cloud, and image embeddings. On the generative front, \acy~employs a multi-modal conditional motion diffusion model with a Transformer backbone to maintain temporal coherence of generated motions, and utilizes autoregressive inpainting to facilitate online motion inference with minimal latency (0.17 seconds). We show that our system provides an effective and robust solution that scales to a diverse dataset of over 200 hours of motion in complex indoor and outdoor environments.

\vspace{-6mm}
\end{abstract}

\section{Introduction}
\label{sec:intro}
Wearable devices such as smart glasses promise to become the cornerstone of next-generation personal computing. A key challenge is accurately interpreting the wearer's motion from the device's limited input signals, taking into account the social and environmental context at the moment. The capability to generate full-body movements solely from a single head-mounted device (HMD) in real-time, outdoors and indoors, will open the door to many downstream applications, including telepresence, fitness and health monitoring, and navigation.

State-of-the-art methods, such as EgoEgo~\cite{egoego23cvpr}, have shown visually impressive results in a similar context. However, these systems operate offline, are optimized for generating short windows of motion, and are mostly trained on a small set of indoor motions. More crucially, they utilize the head-mounted camera only for head pose estimation, missing the opportunity to harness additional image features of the environment and of the wearer's own body.

In this paper, we introduce \acy~(Human Motion Diffusion from HMD), the first system, to our knowledge, capable of online generation of full-body movements from a single HMD (Project Aria Glasses \cite{aria23surreal}), conditioned on outward-facing egocentric camera streams in diverse environments. Given that such devices provide limited observation of the body and surroundings, the critical question is how to maximally utilize the input. Our approach reuses input data to generate features across different modalities, covering independent aspects of the environment and motion. Specifically, from the input streams, we mix and match analytical and learning toolboxes to extract 1. wearer's head motion from off-the-shelf real-time visual SLAM; 2. environment feature points as a by-product of SLAM, important for motion disambiguation in complex scenes; and 3. head camera image embeddings (\eg using CLIP \cite{radford2021learning}) for additional scene clues and intermittently visible body parts. 

However, full recovery of the wearer's motion is still highly under-constrained, given our input. Our system adopts a generative approach with a diffusion-based Transformer backbone to balance motion reconstruction and generation, enabling diverse outcomes, such as varying leg movements, from the same inputs. Additionally, our diffusion model can predict motions with minimal future information (0.17 s), supporting online and real-time use cases.

Contrary to evaluations using large synthetic datasets or small-scale real-world datasets, we train and test our system on the extensive 200-hour real-world Nymeria dataset~\cite{ma2024nymeria} recorded with publicly available head-mounted device, containing various indoor and outdoor activities performed by over 100 subjects with diverse body sizes and demographics. While most existing research on motion tracking is evaluated solely based on reconstruction accuracy, we acknowledge the inherent ambiguity in our problem and evaluate our system on generation fidelity and diversity as well. Our contributions are summarized as follows:

\begin{enumerate}
    \item We present a novel application of online full-body motion generation from a single HMD. The multi-modal feature streams extracted from the device serve as a key ingredient for the system's success across a diverse set of environments.
    \item We employ a multi-modal conditional motion diffusion backbone, effectively balancing between accurate motion reconstruction and the diversity and fidelity of synthesized movements.
    \item We demonstrate the adaption of a time-series motion diffusion model for online autoregressive inference through inpainting, eliminating the dependency on future sensor input and achieving minimal latency.
    \item We evaluate the proposed system with large-scale, real-world Nymeria~\cite{ma2024nymeria} dataset and achieve state-of-the-art performance for single-HMD motion generation.
\end{enumerate}

\section{Related Work}
\label{sec:related}

\PAR{Human Motion from Sparse Sensors.}
Capturing motion with wearable sensors has gained interest across fields like Computer Vision, Graphics, and Health. Self-contained sensors like IMUs \cite{SIP}, electromagnetic sensors \cite{kaufmann2021pose}, and EMGs \cite{chiquier2023muscles} offer motion reconstruction without the need for costly studios with multiple cameras. The sparse sensor placement reduces user friction, but high noise levels require learning methods to improve reconstruction. Examples include six IMUs configurations \cite{SIP,DIP:SIGGRAPHAsia:2018, transpose21tog, tip22siggraph, pip22cvpr}, head and wrists VR trackers \cite{avatarposer22eccv, du2023avatars, zheng2023realistic, castillo2023boDiffusion, divatrack24eg, egoposer23}, and hybrid approaches with an external RGB camera~\cite{yang2022hybridtrak}.

Our approach uses a single wearable device to minimize user friction, though this complicates the recovering of motion. However, for many applications like telepresence, visually appealing, realistic, and diverse inferred motions are often more important than precision. Thus, we evaluate our system not just on reconstruction accuracy but also on realism and diversity -- metrics often overlooked in this field.

\PAR{Pose and Motion from Egocentric Cameras.}
Wearable egocentric cameras are ideal for self-contained motion generation systems, which saw increasing research interest. Two main types of body-mounted cameras -- downward-facing (often fish-eye) and outward-facing -- have been the focus of research. Most studies on downward cameras \cite{rhodin2016egocap, cha2018towards, tome2020selfpose, wang2022estimating, xu2019mo, zhang2021automatic,zhao2021egoglass, park2023domain, kang2023ego3dpose, liu2023egohmr}, directly predict current pose from corresponding images, sacrificing temporal coherence. Wang \etal \cite{wang2023egocentric} addressed this by adopting a diffusion model for temporal regularization in a separate refinement stage, which inspired us to adopt a diffusion backbone and a single-stage time-window-based learning architecture. Both synthesized \cite{unrealego22eccv} and, recently, real-device \cite{hakada2024unrealego2} datasets are used to train and evaluate such methods.

Outward-facing cameras are more common on current devices (e.g. Project Aria \cite{aria23surreal}), though egocentric motion generation is less explored in this setup. A key challenge identified in early work with chest-mounted cameras \cite{jiang2017seeing} is intermittent body visibility, which makes the task underconstrained. Later works \cite{yuan20183d, Yuan_2019_ICCV, luo2021dynamics} explored simulation methods that leveraged physics to address missing motion information. EgoEgo \cite{egoego23cvpr} demonstrated the generalizability of single camera systems to large-scale datasets. We build upon EgoEgo, while utilizing additional visual cues beyond head pose inference and enhancing support for non-flat terrains and low-latency long sequence generation.

There has also been research effort on combining wearable sensors such as IMUs with head-mounted egocentric cameras for accurate motion reconstruction \cite{egolocate23tog, mocapee24lee, hps21cvpr}. Our system can be easily adapted to such multi-device setups as well, which could further improve its accuracy.

\PAR{Learning-based Pose and Motion Generation.}
Generating controllable and realistic human movements is a long-standing goal in computer graphics and vision. Modern deep learning opens new possibilities for this problem, with earlier attempts exploring both regression-based \cite{holden2017phase,holden2020learned} and generative \cite{henter2020moglow, ling2020character} frameworks. Recently diffusion models demonstrated impressive capabilities in the generative setting across various tasks such as text \cite{priormdm24iclr, zhou2023emdm}, music \cite{tseng23cvpr}, and audio \cite{alexanderson2023listen} conditioned motion generation. While the field starts to see conditional diffusion methods where the control signal is temporally dense \cite{castillo2023boDiffusion, du2023avatars}, frameworks that generate motions in an online fashion with minimal latency \cite{van2023Diffusion} are still underexplored. Our work adopts autoregressive inpainting for low-latency inference -- this concept of autoregressive diffusion models has been explored in the motion domain albeit in different contexts \cite{han2023amd, zhang2023tedi, shi2023controllable, yin2023controllable}.

The success of diffusion models in motion synthesis has also intrigued researchers to use them for pose reconstruction, \eg from third-person view, especially when ambiguity exists \cite{choi2023diffupose, ci2023gfpose, foo2023distribution, gong2023diffpose, holmquist2023diffpose, Zhang_Ma_Zhang_Aliakbarian_Cosker_Tang_2023, zhang2024rohm}. Our task is highly ambiguous as well, and our system adopts Transformer-based diffusion models to generate temporally coherent motions.

\PAR{Scene-aware Pose and Motion Modeling.}
Motion generation and reconstruction satisfying scene and environment constraints is critical for learning-based motion models to become practical. Recent work has looked into various methods and representations to incorporate scene information, such as shape primitives \cite{luo2022embodied, lama23iccv}, point-cloud-based networks \cite{huang2023Diffusion, zhao22coins, wang2022towards, wang2022humanise}, voxel-based networks \cite{wang2023scene, hassan2021stochastic, starke2019neural}, scene images \cite{caoHMP2020}, signed distance fields \cite{zhao2023synthesizing}, to name a few. With most methods targeting offline applications and many requiring end-of-motion goal specifications, our scene representation with a per-frame bounding box and autoencoder facilitate online usage and large-scene deployment. The scene points in our method are captured from the same head-mounted device during SLAM without needing additional scanning devices. As a trade-off, the available scene points are sparser and noisier.

\section{Method}
\label{sec:method}
\begin{figure*}[h!]
  \centering
  \vspace{-3mm}
  \includegraphics[width=0.9\textwidth]{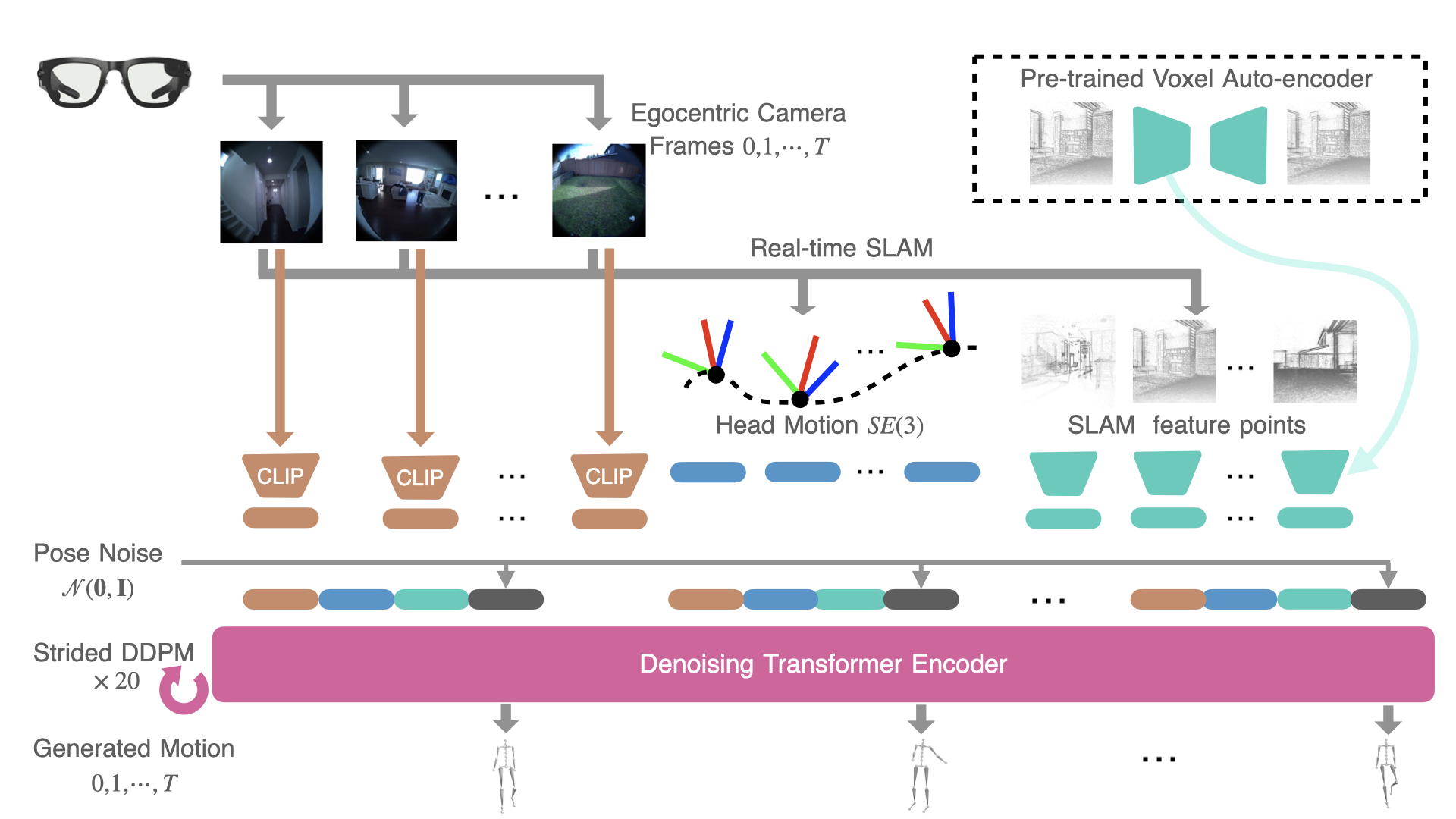}
\caption{Overview: \acy~generates realistic full-body motion that aligns with the signals from a single head-mounted device. Using the image streams from the egocentric camera and head trajectory with the feature cloud from the onboard SLAM system, we employ a diffusion-based framework to generate the wearer's full-body motion.}
  \label{fig:example}
  \vspace{-5mm}
\end{figure*}

We introduce a diffusion-based framework for generating full-body motion based on multi-modal signals from an HMD, like the Project Aria Glasses \cite{aria23surreal}. As shown in Fig.~\ref{fig:example}, our system uses device with an outward-facing camera, capable of real-time SLAM~\cite{mps} (which may utilize other sensors) which produces a 6D pose trajectory, and a spatial map of the environment represented by an aggregated point cloud. We extract contextual information from both the environment point clouds and the egocentric video stream, using a CLIP encoder~\cite{radford2021learning} for image embedding and an independently trained point cloud autoencoder for spatial map embedding to supplement the 6D pose.

Given the under-constrained nature of the task, we employ a diffusion model~\cite{ho2020denoising} with a time-series Transformer encoder~\cite{vaswani2017attention} to model the motion distribution. To ensure temporal consistency during streaming, we use autoregressive inpainting during denoising, aligning new body motion with previous predictions.

\subsection{Multi-modal Scene and Motion Conditions}

Our model is trained to align its output with three modalities of features, all of which are streaming frame by frame to allow infinitely long motion generation. For each frame, the inputs include a head pose $(\bm{t}, \bm{R}) \in \textrm{SE}(3)$ representing the head's position and orientation, a color image $\bm{I}$ from the camera, and a set of SLAM feature points $\bm{S} \in \mathbb{R}^{N \times 3}$ of the surrounding scene. We concatenate features per-frame and process the resulting vector with a linear layer (see supplementary). We elaborate on each modality and their respective design considerations below. 

\PAR{Head Pose Trajectory.} The device pose provides precise spatial location and movement of the wearer's head. We augment the device pose vector with its linear and angular velocity vector $(\bm{v}, \bm{\omega})$ computed from finite differences to form $\bm{p} = \{\bm{t}, \bm{R}, \bm{v}, \bm{\omega}\}$. We canonicalize each window of $\{\bm{p}\}_{0,1, \cdots, T}$ to its first frame $\bm{p}_0$, allowing the model to function in arbitrarily large spaces and generate infinitely long sequences. This is crucial for navigation in a multistory building or outdoor hiking with large elevation changes.

\PAR{Camera Image Embeddings.} Beyond the head pose trajectory derived from SLAM, the egocentric camera images offer additional valuable information. For example, when a body part becomes visible, the image provides a strong cue of the wearer's pose. However, direct utilization of the image content proves less useful, as it may capture distracting texture details when all we need is high-level semantics such as "the left hand is above the waist." Empirically, we found that CLIP embeddings~\cite{radford2021learning}, $E_I(\bm{I})$, provide significant performance boost to the learning process while avoiding overfitting to superficial image characteristics.

\begin{figure}
\centering
  \includegraphics[width=\columnwidth]{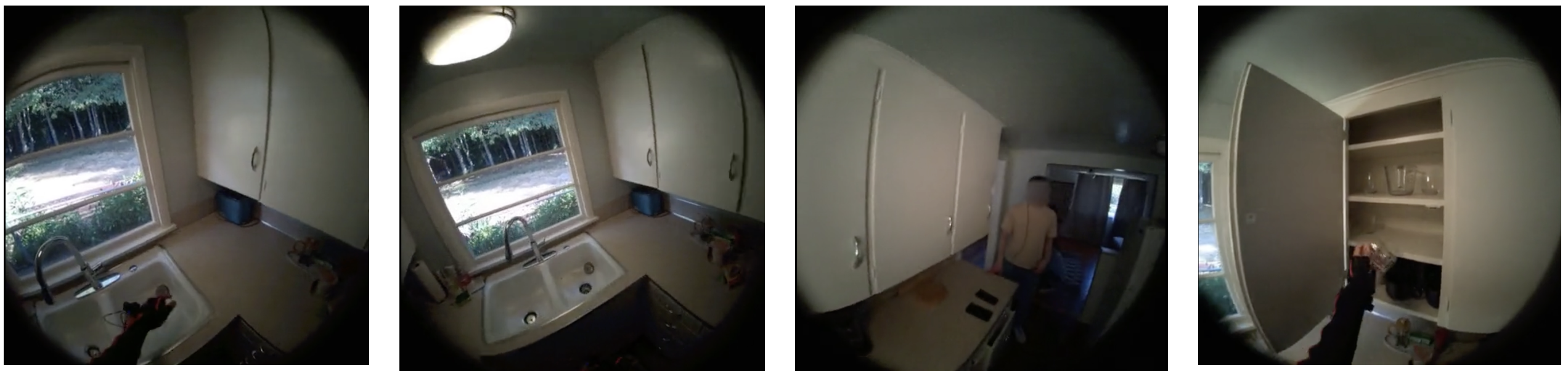}
  \caption{A typical input sequence from egocentric camera with only few body parts of the wearer intermittently visible, rendering standard full-body reconstruction network backbones ineffective.}
    \label{fig:input-example}
  \vspace{-5mm}
\end{figure}

It is crucial to note that embeddings from human-related backbones, such as those trained for pose reconstruction from monocular videos, do not perform well in our case. Figure \ref{fig:input-example} shows a typical input camera sequence when only a few parts of the body (hands in this case) are visible. This differs significantly from downward-facing egocentric cameras, which observe most of the body. This discrepancy leads to failures in existing network backbones for full-body motion, and it may be tempting to assume that such input might not be useful for full-body motion reconstruction. However, high-level descriptions of the images that contain scene information, such as ``hand reaching to the sink (which is typically at a standard height)'' or ``a person kicking a football (implicitly indicating that the wearer might also soon interact with the ball)'', are actually quite useful for spatial reasoning of the wearer's end effectors. We hypothesize that this observation explains why CLIP embeddings are advantageous in our unique problem setting.

\PAR{SLAM Point Cloud Embeddings.} Visual SLAM algorithms identify static feature points in the environment (\eg corners and edges of furniture) and aggregate them over time to build 3D maps. These points offer crucial environment features to constrain motion generation, akin to pre-scanned scenes utilized in prior work~\cite{hps21cvpr, mocapee24lee}. At each frame, we only consider the SLAM feature points $\bm{S}$ within a 2m x 2m x 2m volume. The center of the volume is the current device position offset downwards by one meter, similar to prior works~\cite{starke2019neural}. This ensures the model focuses only on relevant spatial information as the wearer moves around. To better handle the noisy and often incomplete SLAM point clouds, we pre-train an autoencoder on the voxelized SLAM point clouds $V(\bm{S})$ within the bounding volume on all frames in our training dataset and use its encoder $E_S(\cdot)$ to generate point cloud embeddings $E_S(V(\bm{S}))$. While a new map may not offer much information right away, rich point cloud features could quickly build up if the wearer stays in the same environment for a prolonged period (\eg 15 min) or if they have access to a prebuilt map.

\subsection{Conditional Motion Diffusion Model}

Given all input signals from the device, $\bm{c} = \{\bm{p}, E_I(\bm{I}), E_S(V(\bm{S}))\}_{0,1,\cdots,T}$, diffusion models such as DDPM~\cite{ho2020denoising} can model the distribution of all motions conditioned on $\bm{c}$ by progressively introducing distortions (Gaussian diffusion noises) into the motion sequence and learning a neural network model $D$ to reverse these distortions. The sequence of forward distortions can be described by the following equation: 

\begin{small}
\begin{equation}
q(\bm{x}_t | \bm{x}_0, \bm{c}) = \mathcal{N} (\sqrt{\alpha_t} \bm{x}_0, (1-\alpha_t) \mathbb{I}) = \sqrt{\alpha_t} \bm{x}_0 + \sqrt{1-\alpha_t} \bm{\epsilon},
\end{equation}
\end{small}
where the motion $\bm{x} \in \mathbb{R}^{T \times F}$ is represented as a time series with window length $T$ and motion feature dimension denoted as $F$. Here, $\bm{\epsilon} \sim \mathcal{N}(\bm{0}, \mathbb{I})$ denotes the unit Gaussian noise, and $t \in \{0, 1, \cdots, S \}$ signifies the level of distortion, with $t=0$ indicating no distortion and $t=S$ representing maximum distortion such that $\alpha_S=0$ and $\bm{x}_S \sim \mathcal{N}(\bm{0}, \mathbb{I})$. The parameter $\alpha_t$ is a monotonically decreasing scalar that governs the noise schedule. The reverse diffusion process is derived using Bayes' rule and can be expressed as:

\begin{small}
\begin{eqnarray}
&q(\bm{x}_{t-1} | \bm{x}_{t}, \bm{x}_0, \bm{c}) = \mathcal{N}(\sqrt{\alpha_{t-1}}\bm{x}_0 + c_{t}\frac{(\bm{x}_{t} - \sqrt{\alpha_{t}} \bm{x}_0)}{\sqrt{1 - \alpha_{t}}} , \sigma^2_{t}\mathbb{I}), \label{eq:reverse}\\
&c_{t} = \sqrt{1 - \alpha_{t-1} - \sigma^2_{t}},~~~~ \sigma_{t}^2 = (1 - \frac{\alpha_{t}}{\alpha_{t-1}}) \frac{1-\alpha_{t-1}}{1-\alpha_{t}}.
\end{eqnarray}
\end{small}

With $\bm{x}_0$ in Eq. \ref{eq:reverse} estimated by the neural net module $\hat{\bm{x}}_0 = D(\bm{x}_t, \bm{c}, t)$, we can iteratively generate a sequence of samples $(\bm{x}_S, \bm{x}_{S-1}, \ldots, \bm{x}_1, \bm{x}_0)$, initiating from $\bm{x}_S \sim \mathcal{N}(\bm{0}, \mathbb{I})$ and progressing towards the desired motion distribution $q(\bm{x}_0 | \bm{c})$ over $S$ reverse diffusion steps. During model training, we randomly sample $t$ from a uniform distribution $U(0, S)$ for every training data. At inference time, we apply $\bar{S}=20$ evenly spaced strided reverse diffusion steps \cite{nichol2021improved}. 
Note that no Gaussian noise is applied to the condition vector $\bm{c}$. Training loss is defined as:

\begin{equation}
    \mathcal{L} = \mathbb{E}_{\bm{x}_0 \times t \sim U(0, S)} || D(\bm{x}_t, \bm{c}, t) - \bm{x}_0 ||^2,
\end{equation}
We did not find it necessary to include auxiliary loss terms to refine output quality.

\begin{figure}
  \centering
    \includegraphics[width=\columnwidth]{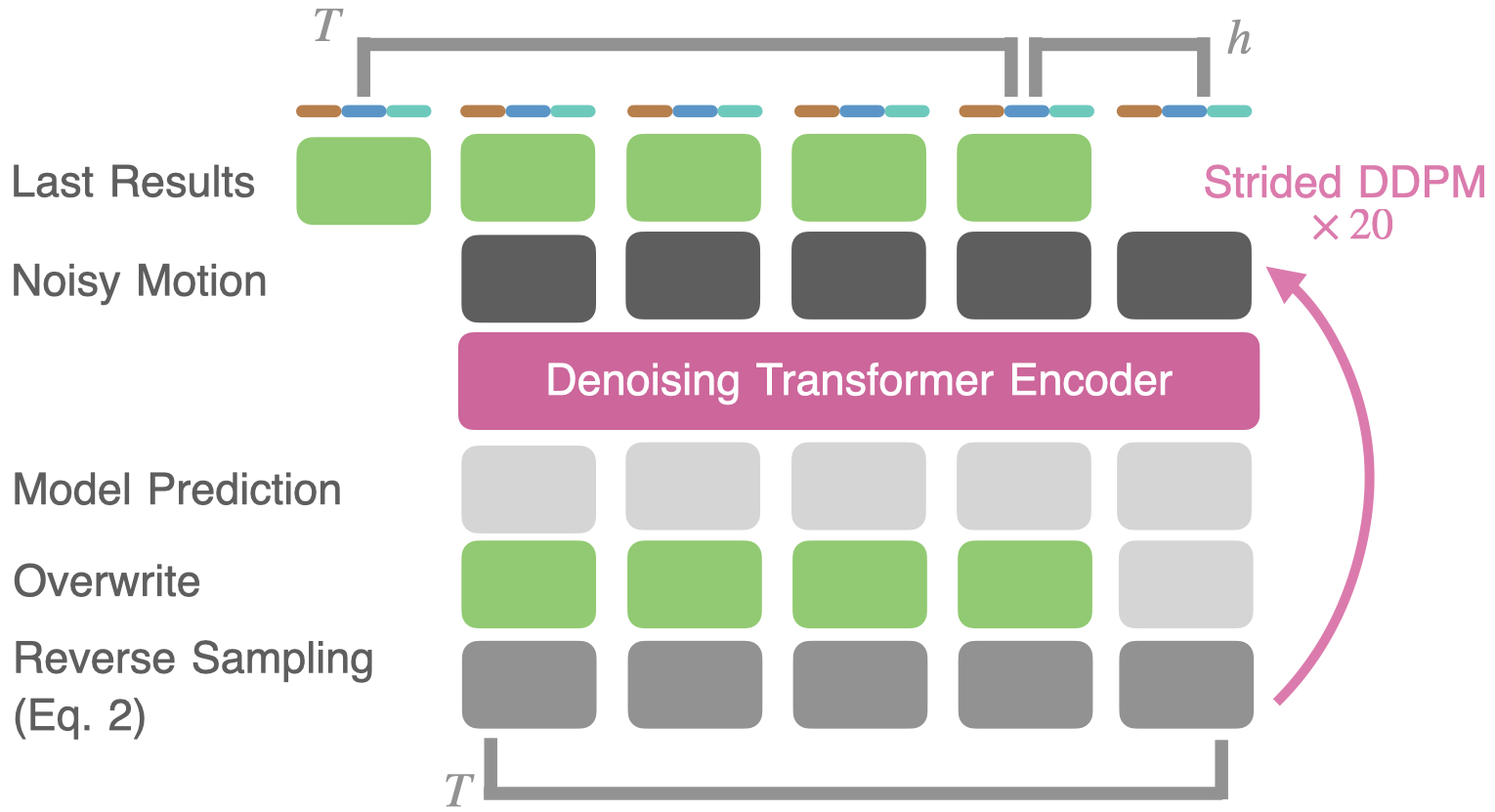}
    \caption{Autoregressive inpainting is performed at each reverse diffusion step to allow long sequence generations both in high- and low-latency settings.}
      \label{fig:method-inpaint}
      \vspace{-5mm}
  \end{figure}

\PAR{Online Inference of Long Sequences.} 
Our motion diffusion model generates up to 4 seconds of motion ($T=240$ frames). To extend this for longer, coherent motions, previous research~\cite{castillo2023boDiffusion, zhang2023diffcollage, priormdm24iclr} suggests generating overlapping windows and enforcing consistency at overlaps during denoising. However, for online generation, we need to remove the dependency on future windows by using an autoregressive approach~\cite{ho2022video}, where each window depends only on the previous one.
Specifically, when two windows overlap by $T-h$ frames (i.e., the current window advances by a stride of $h$), we enforce consistency during each of the $\bar{S}$ denoising steps. After each model evaluation $\hat{\bm{x}}_0 = D(\bm{x}_{t}, \bm{c}, \bm{\tau}_i)$, the prediction $\hat{\bm{x}}_0$ is overwritten by the overlapping prediction from the preceding window:
\begin{equation}
\hat{\bm{x}}_0 = \hat{\bm{x}}_0 \odot \bm{m} + \hat{\bm{x}}_{s0} \odot (1-\bm{m}),
\end{equation}
where $\bm{m} \in  \mathbb{R}^{T \times F}$ is a constant mask that is zero for the initial $T-h$ frames and one for the last $h$ frames. $\hat{\bm{x}}_{s0} = \textrm{cat}(\bm{x}^-_0 [h\textrm{:}T], \bm{0}^{h \times F})$ denotes the prediction from the previous window, shifted by $h$ frames. $\odot$ denotes element-wise multiplication. Following this inpainting operation, we move to denoising step with the updated $\hat{\bm{x}}_0$ using Eq. \ref{eq:reverse}. We report the main results of our system with stride $h = 180$.

However, eliminating the need for future windows is insufficient for online inference with minimal latency since a new window of motion is generated only every $h$ frames, resulting in a latency of $(h-1) \times \delta t$, where $1/\delta t$ is the frame rate. We additionally report our results with $h=10$, indicating a latency of just $0.17$ seconds, close to online requirements. Nonetheless, a smaller $h$ compromises motion quality, as it limits the use of future information. In general, $h$ can be a tunable parameter to trade off quality and latency.

\section{Experiments}
\label{sec:experiments}
We conducted a set of experiments to support these claims:
\begin{itemize}
    \item Our multi-modal conditioning improves motion quality.
    \item Our system achieved high reconstruction accuracy, motion diversity, and physical realism.
    \item Our online (low-latency) variant minimally degrades motion quality compared to high-latency inference.
    \item Our system achieved improved results over state-of-the-art baselines on a large-scale dataset.
    
\end{itemize}
\begin{figure*}[h]
    \small
        \centering
        \includegraphics[width=0.9\linewidth]{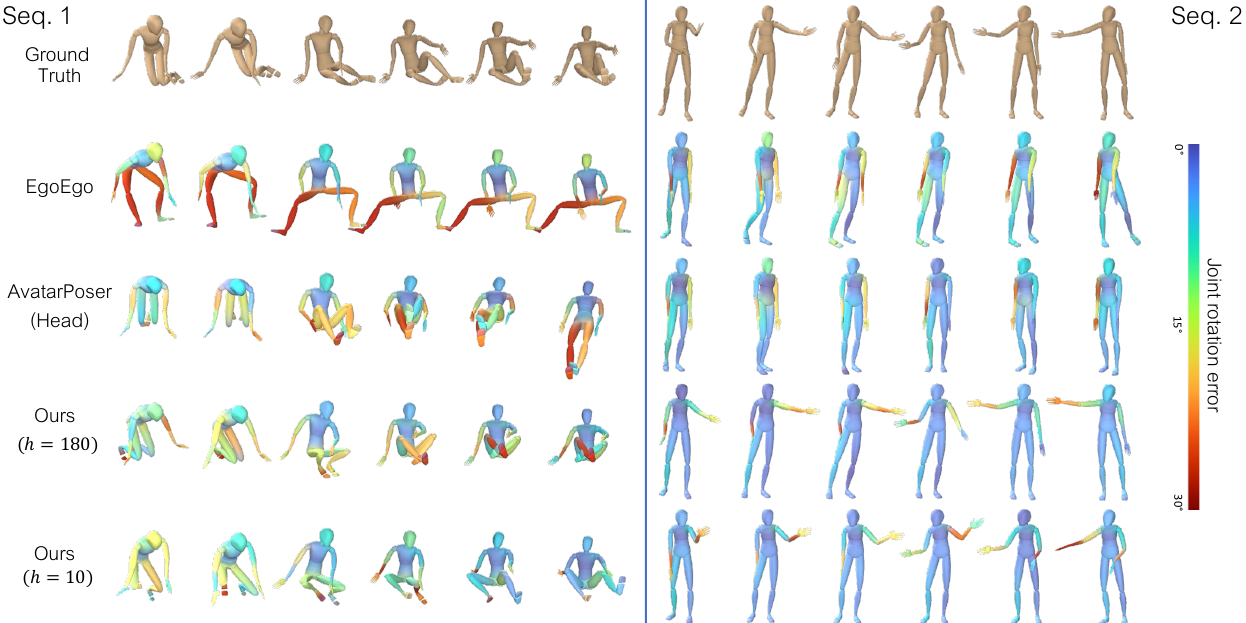}
    
        \caption{Qualitative comparison between \acy~(Ours) and baseline methods.}
        \label{fig:qual_comp}
    
    \end{figure*}
\begin{table*}[ht]
    \centering
    \scriptsize
    \renewcommand{\arraystretch}{1.3}
    \begin{tabular}{l|c|c|c|c|c|c}
    \toprule
    
    & MPJPE $\downarrow$ & Hand PE $\downarrow$ & FID $\downarrow$ & Diversity $\rightarrow$ & Physicality $\rightarrow$ & Floor Pen. $\downarrow$ \\ \hline
    Ground-truth & 0 & 0 & 0 & 16.13 & 0.56  & 0 \\ \hline
    
    EgoEgo & \et{16.61}{1.49} & \et{34.64}{1.64} & \et{35.69}{0.54} & \et{20.15}{0.21} & \et{3.68}{0.74} & \et{2.43}{1.54} \\
    AvatarPoser (Head) & 10.64 & 21.51 & 27.61 & 12.99 & 1.69 & 4.21 \\
    Ours ($h=180$) & \etb{8.36}{0.08} & \etb{16.64}{0.21} & \etb{2.16}{0.02} & \etb{15.74}{0.29} & \etb{1.03}{0.01} & \etb{1.03}{0.06} \\
    Ours ($h=10$) & \et{9.19}{0.05} & \et{17.67}{0.06} & \et{5.00}{0.02} & \et{15.23}{0.02} & \et{1.30}{0.10} & \et{1.19}{0.04} \\
    
    \hline
    \end{tabular}
    \caption{Quantitative results comparing our system with EgoEgo and AvatarPoser.}
    \label{table:main_table}
    \vspace{-0.5cm}
\end{table*}

\PAR{Datasets and Experiment setup.} To address the limitation of evaluating on synthetic or smaller real-world datasets, we train and evaluate our system on a large-scale, first-of-its-kind real-device dataset Nymeria~\cite{ma2024nymeria}. This dataset contains paired multi-modal HMD input signals (captured by Project Aria Glasses~\cite{aria23surreal}) and ground-truth full-body motions (with Xsens inertial motion capture system \cite{xsensawinda}). 
 The dataset covers a diverse range of daily activities and is around 300 hours in size. After initial filtering, we split the data into train, validation, and test split with 202, 3, and 56 hours of data correspondingly. We make sure all subjects and environment scenes in the test split are \textbf{unseen} during training, and distributions of subjects' body sizes and activity scripts are roughly unbiased across the test split. We trained our models with a context window of 240 frames (4 sec) for 20 epochs or 3.5 days with 4 GPUs. The inference is done on a single Nvidia A100 GPU and achieves better than real-time throughput of $>70$ FPS with an online 0.17s-latency ($h=10$) model and $>1350$ FPS with high-latency (3s, $h=180$) model.

\PAR{Baselines.} 
We benchmark our low- and high-latency systems against EgoEgo \cite{egoego23cvpr} and AvatarPoser \cite{avatarposer22eccv}, retraining both models on our dataset. For EgoEgo, we bypass its first stage, using Aria Glasses' SLAM for accurate head motion tracking, and test with its long-sequence inference code. For AvatarPoser, we only provide head motion, masking out wrist device input during training and testing. Unlike the Nymeria paper's short-segment evaluations \cite{ma2024nymeria}, we test all methods with full motion sequences (each around 15min) in an online, autoregressive setting, reflecting real-world use.

\PAR{Metrics.} 
An ideal solution must balance reconstruction accuracy, motion diversity, and physical realism. For instance, when arms are visible to the HMD camera, generated motions should reflect that. When multiple motions are equally valid, \eg sitting, squatting, or kneeling, predictions should cover all possibilities. Finally, any output motion should be visually realistic and within the distribution of physically plausible human movements. We choose metrics that evaluate a system's capability to balance these three goals.

\begin{itemize}
    \item \textbf{Reconstruction:} we report joint position errors (Mean Per Joint Position Errors, MPJPE, in cm) for all methods. As we use the head frame from Aria as the body reference frame for all methods, we assume zero error on head positions or orientations. Instead, we report position errors of the wrist joints (Hand PE, in cm). 
    \item \textbf{Diversity:} Following prior work \cite{raab2023modi,guo2020action2motion}, we report the diversity metric as the mean distance between two same-size randomly sampled subsets from predicted and ground-truth motions in the same latent space as used for FID computation~\cite{guo2020action2motion}.
    \item \textbf{Realism:} we report FID scores measuring the distances in distributions between predicted and ground-truth motions. This is done through training an auto-encoder to construct a motion latent space, following the protocol in Guo \etal 2020 \cite{guo2020action2motion}. We also report the physicality of motions, following the metric proposed in EDGE \cite{tseng2023edge}, which correlates with foot sliding. Lastly, we report the mean floor penetration depth (in cm). Since the floor level varies across time and is non-trivial to estimate for outdoor and complex indoor environments (e.g. the "floor" height for lying in bed should sensibly be the bed height), we adopt a conservative proxy using the lowest joint position of the ground-truth motion across the neighboring 20 seconds.
\end{itemize}

\subsection{Main Results}

We evaluated high- ($h=180$) and low-latency ($h=10$) variants of our system on the 56-hour (224 sequences) test split, averaging 15 minutes per sequence. These test sequences are \textbf{not} cut into short segments to fit the temporal horizon $T$ of the model -- all models are tasked to generate the entire sequence coherently, which is closer to practical application setup. Unlike EgoEgo, where statistics are reported using the best among 200 repetitions, we report the mean and standard deviation of all repetitions. As our test set is very large (e.g. the AMASS \cite{amass19iccv} testing subset used in AvatarPoser contains just two hours of motion), we only run eight repetitions for each of the 224 sequences.

\PAR{Quantitative Results.}
The main quantitative results are summarized in Table~\ref{table:main_table}, with a finer-grained analysis provided in the supplementary. Our system achieved superior performance across all three metric axes of reconstruction, diversity, and realism. As expected, the online variant of our system degrades performance slightly, given inaccessibility of future sensor information, but still outperforms baselines.

Our adapted version of AvatarPoser (referred to as AvatarPoser (Head) in Table~\ref{table:main_table}) performs well, but its frame-by-frame prediction lacks temporal coherence, reducing realism. As a regression model, it captures only the average trend in training data, leading to lower diversity scores. Unlike our multi-modal approach, it lacks environmental awareness, impacting performance (Fig.~\ref{fig:qual_comp}).
EgoEgo generates plausible motions but has two key issues. First, it produces discontinuities during long motion inference, which affect realism metrics. Second, EgoEgo tends to produce overly dynamic arm movements, similar to how some image diffusion models create stylized rather than naturalistic outputs. This leads to higher Hand Position Errors and contributes to increased MPJPE and Diversity scores compared to ground truth. While all the metrics in Table~\ref{table:main_table} are measured as mean across all runs, we additionally report MPJPE of the best-case run: 
8.246, %
14.678, %
for HMD\textsuperscript{2} and EgoEgo (AvatarPoser stays the same).
Compared to Table~\ref{table:main_table}, errors for EgoEgo are noticeably lower but are still behind Ours. 

In summary, our system uniquely balances the accuracy of motion reconstruction and fidelity and diversity of motion generation, surpassing baseline methods. The online variant of our system achieves 0.17-second latency with only a slight degradation in terms of performance, though the gap leaves room for future research and improvement.

\begin{table*}[ht]
    \centering
    \scriptsize
    \renewcommand{\arraystretch}{1.3}
    \begin{tabular}{l|c|c|c|c|c|c}
    \toprule
    
    & MPJPE $\downarrow$ & Hand PE $\downarrow$ & FID $\downarrow$ & Diversity $\rightarrow$ & Physicality $\rightarrow$ & Floor Pen. $\downarrow$ \\ \hline
    Ground-truth & 0 & 0 & 0 & 16.13 & 0.56  & 0 \\ \hline
    Ours, w/o PC, w/o CLIP & \et{9.28}{0.23} & \et{19.47}{0.36} & \et{6.75}{0.08} & \et{14.44}{0.30} & \et{0.90}{0.01} & \et{3.29}{0.31} \\
    Ours, w/ PC, w/o CLIP & \et{8.97}{0.10} & \et{20.38}{0.28} & \et{3.68}{0.03} & \et{15.29}{0.42} & \etb{0.86}{0.00} & \etb{0.99}{0.07} \\
    Ours, w/o PC, w/ CLIP & \et{8.57}{0.11} & \etb{16.32}{0.22} & \et{6.17}{0.02} & \et{14.79}{0.22} & \et{1.01}{0.01} & \et{2.15}{0.15} \\
    Ours, w/ PC, w/ CLIP & \etb{8.36}{0.08} & \et{16.64}{0.21} & \etb{2.16}{0.02} & \etb{15.74}{0.29} & \et{1.03}{0.01} & \et{1.03}{0.06} \\
    \hline
    \end{tabular}
    \caption{\textbf{Ablation study.} \acy~leverages both point cloud (PC) and egocentric video information (CLIP) to reduce per-joint error while keeping the realism and physical plausibility of the motions.}
    \label{table:ablations}
    \vspace{-0.01cm}
    \end{table*}

\begin{figure}
\small
    \centering
    \includegraphics[width=0.95\columnwidth]{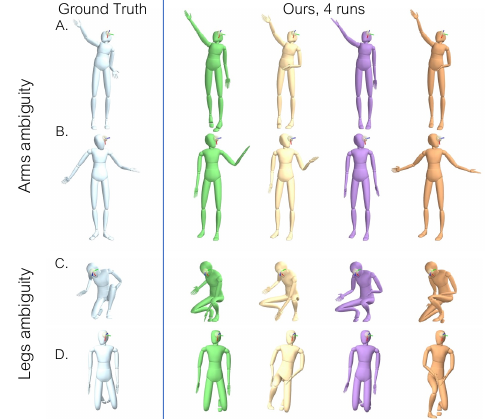}

    \caption{Our system can predict diverse outcomes from identical input (head pose marked as a sphere with coordinate system).}
    \label{fig:diversity}

\end{figure}

\PAR{Qualitative Examples.} Fig.~\ref{fig:qual_comp} visually compares all methods on two motion subsequences from the test set. 
\textit{Sequence 1} shows a complex transition from kneeling to sitting. Regression models like AvatarPoser struggle in under-constrained scenarios, either abruptly switching between poses or averaging them into unnatural ones (e.g., a floating avatar in the last frame). EgoEgo, as a generative model, produces plausible motions but lacks the context to match the ground truth given only head motion.
\textit{Sequence 2} demonstrates another important advantage of our model -- making use of the semantic features from color images. In this ground truth motion, the hands are raised and visible in the camera alternately. We successfully reproduce similar arm movements by conditioning on the CLIP embeddings while both baselines have the arms down.

The generative nature of our model also allows us to produce diverse motions in case of ambiguities. Fig.~\ref{fig:diversity} shows several examples: in the left column, our model generates various plausible states when hands are not visible, such as different poses for the non-visible left hand (seq. A). The right column shows cases with equally possible leg positions, like kneeling vs. squatting (seq. C).

\subsection{Additional Analysis}

\PAR{Ablations.} We ablated our system by removing the point cloud encoder branch (w/o PC) and/or the raw egocentric video branch (w/o CLIP). The results are summarized in Table \ref{table:ablations}, demonstrating the importance of multi-modal scene and motion conditions in our system. 

\begin{figure}
\small
    \centering
    \includegraphics[width=0.9\columnwidth]{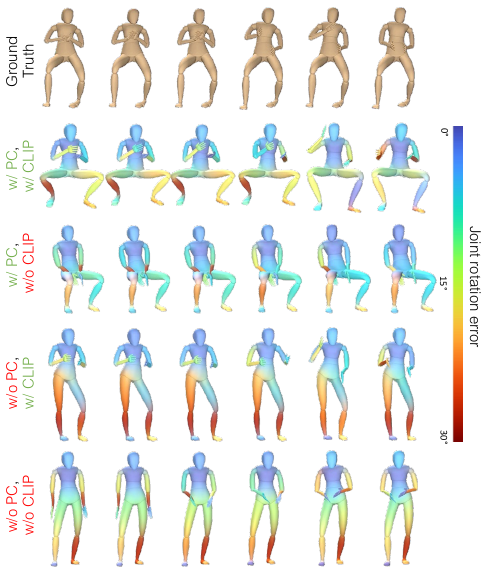}

    \caption{Example motion when ablating the point cloud (PC) or video (CLIP) branches.}
    \label{fig:ablation}

    \vspace{-0.3cm}
\end{figure}

Even without point cloud and CLIP embeddings, our system generates temporally coherent and realistic full-body motions, capturing diverse motion distributions. However, ambiguity arises with head movement alone, such as distinguishing between standing and sitting. Without environmental context, the system might randomly generate or switch between these actions, affecting realism metrics (FID \& Floor Penetration Depth). Table~\ref{table:ablations} shows that point cloud embeddings help align motions with ground truth and reduce environment interpenetration, improving realism. The image encoder also enhances reconstruction accuracy by using semantic clues, particularly when hands are visible. This reduces MPJPE by encouraging specific poses, however it also mildly affects the realism of motion, hence Physicality metric slightly degrades. Fig.~\ref{fig:ablation} illustrates that PC embeddings enable correct sitting motion detection, while image embeddings improve hand motion accuracy. Together, they produce more accurate and realistic results.

\PAR{Error Distribution.} 
As we evaluate on a large scale dataset of realistic daily activities, the metric statistics could be skewed and dominated by mundane actions such as sitting or standing still, or walking from A to B. The more interesting and challenging scenarios that highlight core issues may fall into a long-tail distribution and be obscured by the mean error. To this end, we also report the top 5\% errors in Table~\ref{table:q95}, which is more representative of improvements we expect from our approach.

\begin{table}[ht]
\centering
\scriptsize
\renewcommand{\arraystretch}{1.3}
\begin{tabular}{l|c|c|c}
\toprule

& MPJPE $\downarrow$ & Hand PE $\downarrow$ & Floor Pen. $\downarrow$ \\ \hline
Ground-truth & 0 & 0  & 0 \\ \hline

Ours, w/o PC, w/o CLIP & \et{18.31}{0.89} & \et{40.15}{1.17} & \et{12.91}{1.75} \\
Ours, w PC, w/o CLIP & \et{16.65}{0.44} & \et{41.68}{1.05} & \etb{3.97}{0.32} \\
Ours, w/o PC, w/ CLIP & \ets{16.30}{0.55} & \etb{34.25}{0.90} & \et{8.28}{0.78} \\
Ours, w/ PC, w/ CLIP & \etb{15.49}{0.38} & \ets{34.86}{0.92} & \ets{4.22}{0.28} \\
\hline
\end{tabular}
\caption{Ablation study on the top 5\% per-frame errors (95\% performance), showing significant reduction of peak errors by our multi-modal conditioning.}
\label{table:q95} 
\vspace{-0.4cm}
\end{table}

\section{Conclusions and Discussions}
\label{sec:conclusion}

We introduced \acy{}, a diffusion-based framework for online motion generation from a single head-mounted device. By combining camera-based image embeddings with SLAM-derived head trajectories and semi-dense point clouds, \acy{} produces diverse, natural motions aligned with the environment. Our evaluation across various settings and activities shows that \acy{} outperforms state-of-the-art methods in accuracy, diversity, and realism.

Our insight of leveraging egocentric image features and the capability of modern SLAM systems opens up many new opportunities. For instance, we could incorporate more comprehensive contextual information available from recent advancements in image understanding, including depth estimation from monocular videos, panoptic segmentation, or scene reconstruction through neural radiance fields or 3D Gaussian Splats. Additionally, we envision leveraging video embeddings over extended context windows, potentially from visual language models (VLMs)~\cite{achiam2023gpt}, to refine context conditions further. 

Currently, the performance of our system is still limited by available context information. For example, the CLIP embeddings cannot provide precise spatial information, so they fall short of constraining the precise pose of the hands even when they are visible. The noisy and sparse point clouds are less ideal than dense depth maps for accurate environment contact information; the errors from the SLAM reconstruction can also propagate to our system. On the other hand, incorporating more and denser input streams pose challenge in runtime performance. We will further elaborate on the above limitations in the supplementary.

{\footnotesize
{
\PAR{Acknowledgments:} We thank the Surreal team, especially Svetoslav Kolev, Hyo Jin Kim, Rowan Postyeni, and Renzo De Nardi, for their valuable discussions and help in the project. Gerard Pons-Moll is supported by the German Federal Ministry of Education and Research (BMBF): Tübingen AI Center, FKZ: 01IS18039A, by the Deutsche Forschungsgemeinschaft (DFG, German Research Foundation) – 409792180 (Emmy Noether Programme, project: Real Virtual Humans). Gerard Pons-Moll is a member of the Machine Learning Cluster of Excellence, EXC number 2064/1 – Project number 390727645 and is supported by the Carl Zeiss Foundation. Yifeng Jiang is partially supported by the Wu Tsai Human Performance Alliance at Stanford University.
}
}

\clearpage

\renewcommand{\thesection}{\Alph{section}} 
\setcounter{section}{0}
\twocolumn[{
\renewcommand\twocolumn[1][]{#1}
\newpage
\null
\vskip .375in
\begin{center}
  {\Large \bf SUPPLEMENTARY MATERIALS \\ \acy: Environment-aware Motion Generation from Single Egocentric Head-Mounted Device \par}
  \vspace*{24pt}
\end{center}
}]

\section{Technical details}
\label{sec:supp_method}
\PAR{Architecture and motion inference.}

Our conditional motion diffusion model follows the Transformer-based architectures presented in EDGE~\cite{tseng2023edge} and DiT~\cite{peebles2023scalable} with additional MLP encoder layers to gradually reduce the input dimension (which is bigger due to added CLIP and PC features) to the token latent space size. Our input consists of the motion input (as a translation, rotation, and linear and angular velocities) and PC and CLIP features, all concatenated together, representing one sequence token per frame. Following AvatarPoser~\cite{avatarposer22eccv}, the model only predicts local joint rotations but not global translation. The global movement of the character is created during test time by ``stitching'' the predicted body motion to the ground-truth head motion, and the head motion can be directly obtained through real HMD motion obtained through SLAM, offset by a constant calibration matrix provided by the dataset. The motion output of the diffusion model is denoted as $\bm{x} \in \mathbb{R}^{T \times F}$, where $T=240$ and $F=23\times6$. The skeleton following Xsens definition has 23 ball-and-socket joints, and for each joint, the output rotation is represented as the first two columns of its local rotation matrix. Note that the definition of Xsens human skeleton is very similar to SMPL \cite{smpl2015loper}, with the main difference being the ordering of joints. The model is not conditioned on body size information, but during training, it is forced to see HMD input motions from different subjects covering highly diverse demographics. As such, the trained model is able to handle body size variation implicitly. However, providing size information as an explicit condition might further improve model performance and reduce visual artifacts such as floor penetration and foot sliding. To create the motion visualizations and compute position error metrics, we used ground truth body sizes (skeleton bone lengths) for each subject.

\PAR{Image encoder.} We use CLIP~\cite{radford2021learning} variation ViT-L/14 for our experiments and compute embeddings from the timestamp-synchronized 30 FPS camera; to get the 60 FPS image feature condition, we duplicate every frame one more time.
We also tried other image encoders and found that CLIP features perform best for our task -- please refer to Sec.~\ref{suppsec:image_encoders} for experimental results.

\begin{table*}[ht]
\centering
\scriptsize
\renewcommand{\arraystretch}{1.3}
\begin{tabular}{l|c|c|c|c|c|c}
\toprule

& MPJPE $\downarrow$ & Hand PE $\downarrow$ & FID $\downarrow$ & Diversity $\rightarrow$ & Physicality $\rightarrow$ & Floor Pen. $\downarrow$ \\ \hline
Ground-truth & 0 & 0 & 0 & 16.13 & 0.56  & 0 \\ \hline

Ours w/ DINOv2 & \et{8.72}{0.07} & \et{17.24}{0.18} & \et{2.45}{0.02} & \et{15.38}{0.19} & \etb{0.91}{0.00} & \et{1.42}{0.07} \\
Ours w/ VC-1 & \et{8.54}{0.11} & \etb{16.64}{0.22} & \et{4.34}{0.06} & \et{15.00}{0.42} & \et{0.92}{0.01} & \et{1.26}{0.10} \\
Ours w/ CLIP (current) & \etb{8.36}{0.08} & \etb{16.64}{0.21} & \etb{2.16}{0.02} & \etb{15.74}{0.29} & \et{1.03}{0.01} & \etb{1.03}{0.06} \\

\hline
\end{tabular}
\caption{Comparison between different image feature encoders. MPJPE, Hand PE and Floor penetration are in cm.}
\label{table:imgenc}
\end{table*}

\PAR{Pointcloud encoder.}
As mentioned in the main paper, the pointcloud encoder considers only SLAM points within the 2m x 2m x 2m volume centered around the head with 1m offset downward. The points are voxelized in a 10x10x10 voxel grid in the following way: for each voxel center, the closest point is selected and the distance is stored as a voxel value. All the distances are truncated at 10cm (so the value is clipped between 0 and 0.1). The voxel volume is rotated with the head orientation but only along the Z (gravity) axis. 

The PC autoencoder consists of the encoder and decoder parts; the encoder consists of 4 convolution layers with $3\times3$ kernel, channel sizes 16,32,64,128 correspondingly, ReLU in between, with the average pooling in the end to produce one feature vector of size 128. Decoder is an inversion of that, consisting of 4 transposed convolution layers. It is trained on the volumes extracted using our train set's point clouds and head trajectories. We train with Adam~\cite{adam15kingma} optimizer and learning rate of $10^{-3}$ for 10 epochs.

\PAR{System runtime.}
Our current implementation assumes that point cloud encodings and CLIP features are precomputed or computed in parallel on a separate device. The performance will be affected if all computations need to happen on the same device. However, we observed that even in this situation, we could achieve a throughput of $\sim 61$ FPS for our low-latency variant, therefore keeping up with real-time speed: CLIP embeddings take around 5 ms to compute per image (2.5 ms per motion frame since we are duplicating every frame), and point cloud encoder taking around 0.1 ms per motion frame. Note that the runtime performance is evaluated on a powerful GPU, which indicates a gap for our system to work in real-time on board of the HMD itself. Additionally, our current implementation assumes the access of all SLAM feature points in the around 15min window of the whole motion sequence. In a true real-time setting, this simplification would require a warm-up phase in the same environment of similar time length.

\section{Dataset details}
\label{sec:supp_dataset}
The Nymeria dataset we used \cite{ma2024nymeria} is captured from Project Aria glasses~\cite{Aria} paired with XSens~\cite{xsens} IMU motion capture suit. The Project Aria glasses are set to record 30fps color video at 1408$\times$1408 pixel resolution. Data captured from the glasses are further processed with its machine perception service (MPS)~\cite{aria23surreal} to output the head transformation and point clouds. The XSens motion data is recorded onboard at 1KHz and processed with Analyse Pro as 240Hz full-body motion, downsampled to 60Hz for our input. The body motion from XSens is synchronized with Aria data to high accuracy using a custom timecode device. The body motion is further calibrated to the Aria head transformation to reduce spatial drift.

The full dataset contains 1200 motion sequences totaling 300 hours of daily activities of 264 participants across 50 locations, from which we used 1040 due to spatial synchronization problems in some sequences. Participants are recruited to cover uniform demographics along the axes of gender, age, height, and weight. The locations include 47 AirBnbs, where 31 are multi-floor houses. There is also a cafeteria with an outdoor patio, a multistory office building, and a campus with a parking lot and multiple biking/hiking trails.

The dataset covers a wide range of daily activities. The highest occurrences are cooking (13.5\%), searching objects (11.0\%), free-form activity improvise (10.4\%), and playing games (10.1\%), whereas the lowest occurrences include working at a desk (1.6\%), locomotion (2.2\%), activities in the office (2.3\%), and creating a messy home (2.3\%). Outdoor activities consist approximately 15\% of the data. For additional details of the dataset, we refer readers to the Nymeria paper \cite{ma2024nymeria}.

We split the dataset for training/validation/testing as 806/10/224 sequences, corresponding to 202/3/56 hours. \textbf{The testing split does not contain any locations or subjects that appear in the training set} to ensure no data leakage. We also strive to maintain a similar distribution of activities between the training set and the test set.

\section{Additional experiments}
\label{sec:supp_experiments}
\begin{table*}[ht]
\centering
\scriptsize
\renewcommand{\arraystretch}{1.3}
\begin{tabular}{l|c|c|c|c|c|c}
\toprule

& MPJPE $\downarrow$ & Hand PE $\downarrow$ & FID $\downarrow$ & Diversity $\rightarrow$ & Physicality $\rightarrow$ & Floor Pen. $\downarrow$ \\ \hline
Ground-truth & 0 & 0 & 0 & 16.95 & 0.04  & 0 \\ \hline

$h=230$ & \et{9.53}{0.01} & \et{16.15}{0.04} & \etb{13.44}{0.01} & \et{15.28}{0.01} & \et{0.32}{0.00} & \et{1.47}{0.02} \\
$h=220$ & \et{9.49}{0.02} & \et{16.07}{0.06} & \et{13.61}{0.01} & \et{15.30}{0.01} & \et{0.25}{0.00} & \et{1.46}{0.01} \\
$h=200$ & \et{9.44}{0.01} & \etb{16.03}{0.04} & \et{13.74}{0.01} & \et{15.32}{0.01} & \et{0.23}{0.00} & \et{1.45}{0.02} \\
$h=180$ (Ours) & \etb{9.42}{0.02} & \et{16.05}{0.02} & \et{13.76}{0.01} & \et{15.43}{0.01} & \etb{0.22}{0.00} & \et{1.44}{0.01} \\
$h=120$ & \et{9.43}{0.03} & \et{16.05}{0.05} & \et{14.02}{0.01} & \et{15.22}{0.01} & \et{0.26}{0.00} & \et{1.43}{0.01} \\
$h=60$ & \et{9.49}{0.06} & \et{16.19}{0.03} & \et{14.23}{0.01} & \et{15.20}{0.01} & \et{0.30}{0.00} & \et{1.33}{0.03} \\
$h=30$ & \et{9.61}{0.04} & \et{16.42}{0.07} & \et{14.39}{0.03} & \et{15.57}{0.03} & \et{0.40}{0.00} & \et{1.26}{0.03} \\
$h=20$ & \et{9.75}{0.10} & \et{16.51}{0.08} & \et{16.46}{0.04} & \et{15.36}{0.04} & \et{0.45}{0.00} & \etb{1.18}{0.05} \\
$h=10$ (Ours low-lat.) & \et{10.19}{0.12} & \et{17.13}{0.14} & \et{17.00}{0.10} & \et{15.66}{0.10} & \et{0.73}{0.03} & \et{1.41}{0.14} \\
$h=5$ & \et{13.13}{0.46} & \et{21.28}{0.45} & \et{20.36}{0.33} & \etb{16.71}{0.33} & \et{0.94}{0.02} & \et{1.84}{0.43} \\
$h=3$ & \et{21.10}{1.08} & \et{29.80}{1.15} & \et{72.63}{0.82} & \et{20.35}{0.82} & \et{1.29}{0.12} & \et{4.49}{0.51} \\
$h=1$ & \et{28.96}{1.68} & \et{38.13}{1.54} & \et{129.94}{1.37} & \et{22.74}{1.37} & \et{2.22}{0.17} & \et{3.75}{0.72} \\

\hline
\end{tabular}
\caption{Ablation study on the latency ($h$) parameter. Test is performed on a subset (9\%) of the current test split. MPJPE, Hand PE and Floor penetration are in cm.}
\label{table:stride_ablation}
\end{table*}

\begin{table*}[ht]
\centering
\scriptsize
\renewcommand{\arraystretch}{1.3}
\begin{tabular}{l|c|c|c|c|c|c}
\toprule

& MPJPE $\downarrow$ & Hand PE $\downarrow$ & FID $\downarrow$ & Diversity $\rightarrow$ & Physicality $\rightarrow$ & Floor Pen. $\downarrow$ \\ \hline
Ground-truth & 0 & 0 & 0 & 16.95 & 0.04  & 0 \\ \hline

2 steps & \et{9.54}{0.01} & \et{15.94}{0.02} & \et{15.04}{0.00} & \et{15.45}{0.00} & \et{0.50}{0.00} & \et{1.87}{0.02} \\
3 steps & \et{9.27}{0.01} & \etb{15.52}{0.03} & \et{15.28}{0.01} & \et{14.85}{0.01} & \et{0.32}{0.00} & \et{1.64}{0.01} \\
5 steps & \etb{9.26}{0.01} & \et{15.57}{0.03} & \et{14.94}{0.01} & \et{14.97}{0.01} & \et{0.25}{0.00} & \et{1.54}{0.02} \\
10 steps & \et{9.34}{0.02} & \et{15.81}{0.03} & \et{14.25}{0.01} & \et{15.50}{0.01} & \et{0.24}{0.00} & \et{1.47}{0.01} \\
20 steps (Ours) & \et{9.42}{0.02} & \et{16.05}{0.02} & \et{13.76}{0.01} & \et{15.43}{0.01} & \etb{0.22}{0.00} & \et{1.44}{0.01} \\
40 steps & \et{9.52}{0.02} & \et{16.21}{0.02} & \et{13.40}{0.01} & \et{15.71}{0.01} & \et{0.22}{0.00} & \et{1.43}{0.02} \\
80 steps & \et{9.60}{0.03} & \et{16.38}{0.02} & \etb{13.11}{0.01} & \etb{15.77}{0.01} & \et{0.23}{0.00} & \etb{1.41}{0.01} \\

\hline
\end{tabular}
\vspace{0.1cm}
\caption{Ablation study on the amount of steps in reverse diffusion process. Test is performed on a subset (10\%) of the current test split.}
\label{table:diffstep_ablation}
\end{table*}

\subsection{Metrics -- units of measure and symbols}
All the metrics shown here and in the main paper, that have units of measure, namely positional errors (MPJPE, Hand PE, Low. PE, Up. PE) and Floor Penetration, are presented in cm. The down arrow $\downarrow$ means that lower value is always better for this metric, and the right arrow $\rightarrow$ means that the value closer to Ground-truth is better.

\subsection{Comparison between different images feature encoders}
\label{suppsec:image_encoders}
To explain our choice of CLIP~\cite{radford2021learning} feature as a feature encoder, we additionally trained two versions of our method with image features produced by DINOv2~\cite{oquab2023dinov2} and VC-1~\cite{majumdar2024we} feature encoders. For VC-1, we chose the best performing ViT-L model, with embedding size of 1024 and input size of $250\times250$ (cropped to $224\times224$ during preprocessing); for DINOv2, we chose second to largest model ViT-L/14, providing it with the input of the same size (padded to $252\times252$) and taking the class token of the output (size 1024), which corresponds to the global image description as it gathers the information from all the image patches.
The comparison is presented in Tab.~\ref{table:imgenc}. We found that, while methods VC-1 and DINOv2 have close generation precision and a slight advantage in Physicality (correlated to foot sliding), the model with CLIP features shows the best results on most metrics, proving our choice of the image feature encoder.

\begin{table*}[ht]
\centering
\scriptsize
\renewcommand{\arraystretch}{1.3}
\begin{tabular}{l|c|c|c|c|c}
\toprule

& MPJPE $\downarrow$ & Hand PE $\downarrow$ & Low. PE $\downarrow$ & Up. PE $\downarrow$ & Floor Pen. $\downarrow$ \\ \hline
EgoEgo & \et{16.61}{1.49} & \et{34.64}{1.64} & \et{26.58}{3.57} & \et{11.31}{0.54} & \et{2.43}{1.54} \\
AvatarPoser (Head) & 10.64 & 21.51 & 17.70 & 6.90 & 2.94 \\
AvatarPoser  (Head \& Hands) & \bf7.74 & \bf6.29 & 16.10 & \bf3.11 & 4.63 \\
Ours, w/o PC, w/o CLIP & \et{9.28}{0.23} & \et{19.47}{0.36} & \et{15.04}{0.53} & \et{6.21}{0.11} & \et{3.29}{0.31} \\
Ours, w/ PC, w/o CLIP & \et{8.97}{0.10} & \et{20.38}{0.28} & \ets{13.59}{0.21} & \et{6.53}{0.07} & \etb{0.99}{0.07} \\
Ours, w/o PC, w/ CLIP & \et{8.57}{0.11} & \ets{16.32}{0.22} & \et{14.02}{0.25} & \ets{5.64}{0.06} & \et{2.15}{0.15} \\
Ours, w/ PC, w/ CLIP & \ets{8.36}{0.08} & \et{16.64}{0.21} & \etb{13.23}{0.16} & \et{5.75}{0.06} & \ets{1.03}{0.06} \\

\hline
\end{tabular}
\caption{Lower and upper body error depending on the input variations. We are beating a 3-point input baseline on a lower body error and achieve close performance on average. All the metrics are in cm.}
\label{table:main_upper_lower}
\vspace{-0.2cm}
\end{table*}

\begin{table*}[ht]
\centering
\scriptsize
\renewcommand{\arraystretch}{1.3}
\begin{tabular}{l|c|c|c|c|c}
\toprule

& MPJPE $\downarrow$ & Hand PE $\downarrow$ & Low. PE $\downarrow$ & Up. PE $\downarrow$ & Floor Pen. $\downarrow$ \\ \hline
EgoEgo & \et{30.91}{4.82} & \et{60.81}{2.98} & \et{58.63}{12.17} & \et{19.26}{1.16} & \et{10.33}{5.90} \\
AvatarPoser (Head) & 22.09 & 43.19 & 44.18 & 13.01 & 18.96 \\
AvatarPoser (Head \& Hands) & 16.48 & \textbf{11.23} & 37.91 & \textbf{5.63} & 18.15 \\
Ours, w/o PC, w/o CLIP & \et{18.31}{0.89} & \et{40.15}{1.17} & \et{34.35}{2.20} & \et{11.75}{0.37} & \et{12.91}{1.75} \\
Ours, w/ PC, w/o CLIP & \et{16.65}{0.44} & \et{41.68}{1.05} & \et{28.72}{1.02} & \et{12.29}{0.30} & \etb{3.97}{0.32} \\
Ours, w/o PC, w/ CLIP & \et{16.30}{0.55} & \et{34.25}{0.90} & \et{29.98}{1.35} & \et{10.58}{0.26} & \et{8.28}{0.78} \\
Ours, w/ PC, w/ CLIP & \etb{15.49}{0.38} & \et{34.86}{0.92} & \etb{27.35}{0.81} & \et{10.80}{0.26} & \et{4.22}{0.28} \\
\hline
\end{tabular}
\caption{Lower and upper body error study on top 5\% errors (mean of 95\% percentiles across all sequences). Here, we are beating 3-point error baseline on mean per-joint positional error. All the metrics are in cm.}
\label{table:q95_more}
\vspace{-0.2cm}
\end{table*}

\subsection{Ablation study on $h$ parameter values and diffusion steps}

In Tab.~\ref{table:stride_ablation}, we show how the error metrics change depending on the latency ($h$) parameter. Because experiments with $h=$ 1, 3, and 5 take a long time to process on our large test split, we performed this ablation on a 9\% (20 out of 224 sequences) subset of test data. To keep the subset informative and maintain the diversity of activities, we picked one random sequence from each activity scenario. The results in the table demonstrate that the top performance in terms of MPJPE is achieved at $h=180$, which we chose as our default value. While it is not the best on all the metrics, the difference is not as significant. Our low-latency method ($h=10$) demonstrates some performance drop, but not as big compared to the next value $h=5$, keeping a balance between the quality and the output lag.

We also measured metrics change w.r.t. the amount of diffusion steps we taking during inference. Tab.~\ref{table:diffstep_ablation} shows that FID score increases with the amount of steps -- visually, this corresponds to less jittery and more realistic motion. However, the precision of the motion, measured by MPJPE metric, peaks at 5 steps for full body and 3 steps for hands. Therefore, our choice of 20 steps is a balance between motion precision and realism.

\subsection{More results on error distribution}

In Tab.~\ref{table:main_upper_lower}, we present additional metrics, splitting per-joint average error into average error across upper (Up. PE) and lower (Low. PE) body regions. The upper region is defined as all the joints that are higher than the pelvis for the subject standing in a T-pose, namely the spine, shoulders, arms, hands, neck, and head. The lower body region is defined as the rest of the joints, excluding the root joint (hips, legs, feet). From these metrics, we can directly observe the effect of adding pointcloud and image encoders to our data. When the PC encoder is added, the lower body error is reduced significantly, and the upper body gets slightly worse (most likely due to noisy points near the upper body region). This suggests that pointcloud helps to disambiguate the lower body by providing landscape information (floor level, nearby objects, etc.). On the other hand, when CLIP image encoding is added, we notice a major reduction in the upper body error, suggesting that image features help the method better understand interactions and localize hands. At the same time, lower body error also decreases - most likely, the error is reduced when parts of the lower body are visible on camera. HMD$^2$, denoted as ``Ours, w/ PC, w/ CLIP'' in the table, combines both strengths of the methods above and achieves the lowest mean per-joint error. 

\subsection{More top 5\% error results and metric computation algorithm}
The error reduction effect discussed above can also be noted in Tab.~\ref{table:q95_more}, showing the top 5\% error for upper and lower body error metrics. Here, we want to clarify our top error selection strategy. As shown in Sec.~\ref{suppsec:errvar} and Fig.~\ref{suppfig:mpjpe_by_script}, the average error on the sequence greatly depends on the activity performed in that sequence. If we were to sort all the per-frame joint errors and select the top 5\% (95\% percentile) among them, we would only select the frames from several worse-performing sequences. To avoid such behavior, we compute the 95\% error percentile within each sequence separately and average those results across all sequences.

\subsection{Effects of the input variation on the generation performance}
In Tab.~\ref{table:main_upper_lower}, we also present a study of another, much more challenging baseline -- a 3-point input method. For that, we chose the original implementation AvatarPoser~\cite{avatarposer22eccv}, which takes not only the head position and orientation as an input but also the positions and orientations of the hands. With more input information, this baseline achieves better performance on average. However, we highlight that even with additional motion input, it is worse than Ours at generating lower body motion, as Lower body PE is higher. It is important to note that HMD$^2$ achieves \textit{best performance} on the most challenging frames of the sequences even when compared to a 3-point input baseline, as shown in the top 5\% error study in Tab.~\ref{table:q95_more}.

\subsection{Diversity of results given the same input}
\begin{figure*}
  \centering
  \includegraphics[width=0.495\textwidth]{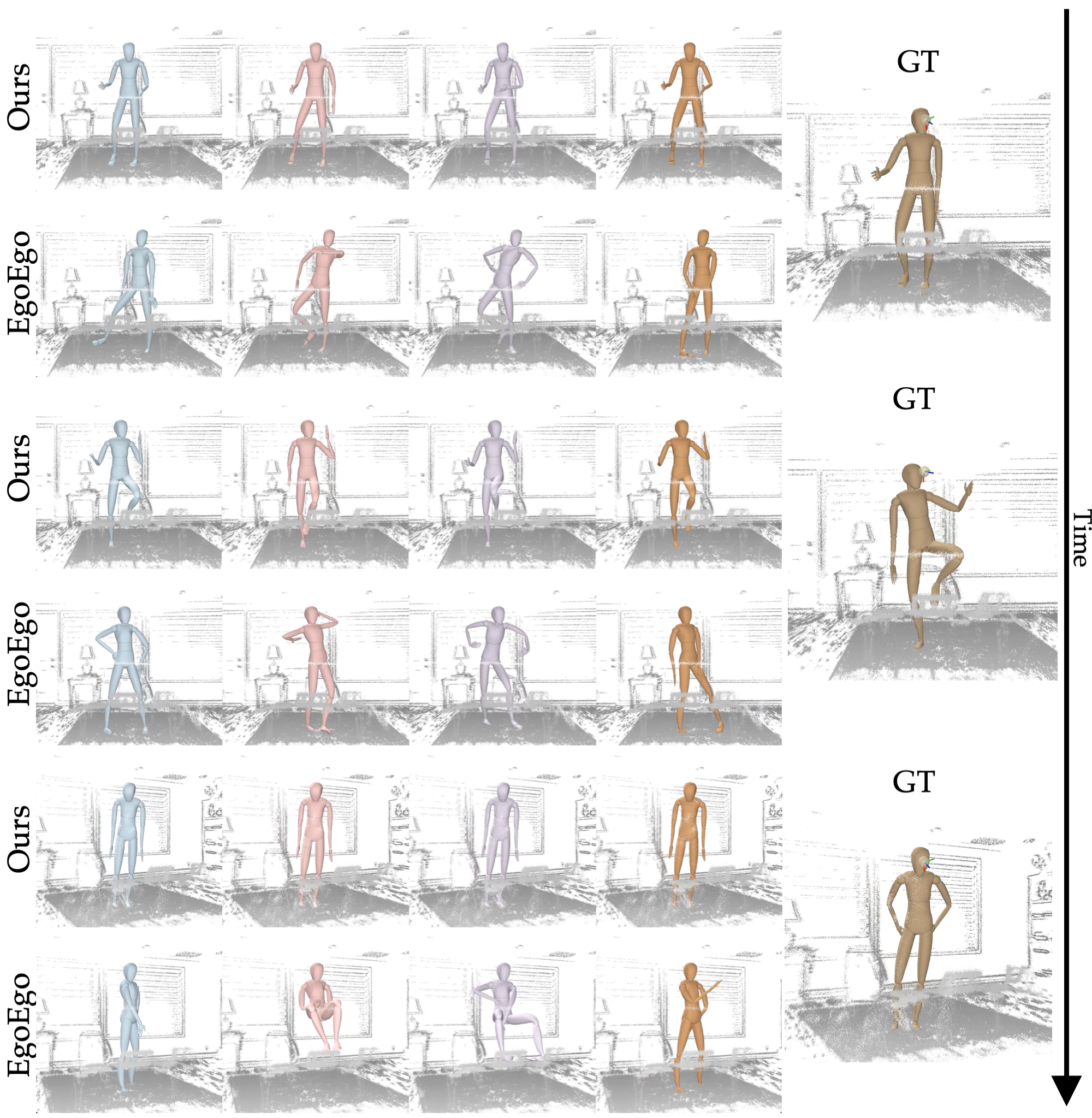}
  \includegraphics[width=0.495\textwidth]{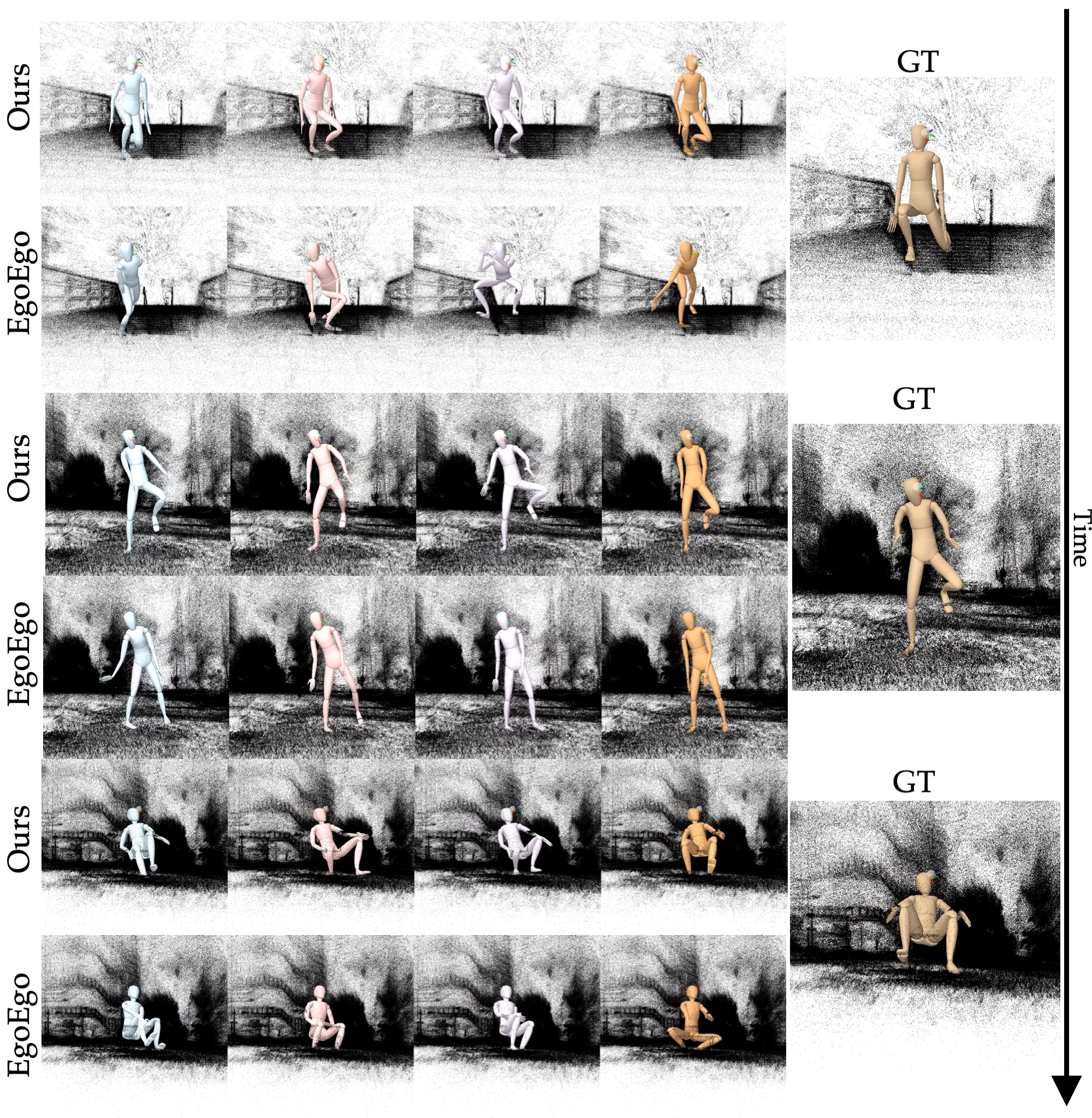}
\caption{Range of possible results given the same input for HMD\textsuperscript{2} and EgoEgo. Colors denote different runs, sequence frame time is increasing from top to bottom.}
  \label{suppfig:diversity}
\end{figure*}

Fig.~\ref{suppfig:diversity} shows 4 random motion samples given the same input for two sequences (1st sequence indoor, 2nd sequence outdoor). A few observations worth highlighting:
1. EgoEgo is also capable of generating diverse predictions, sometimes more diverse than Ours;
2. However, EgoEgo generations tend to be of lower quality - possibly due to model architecture not being as scalable to a massive dataset as Ours and autoregressive long sequence inference not working as well;
3. Moreover, EgoEgo samples often do not satisfy floor height constraints (1st seq. 3rd frame; 2nd seq. 1st frame), and cannot utilize image observation when certain body parts are visible (1st seq., see the right arm in 1st frame and left arm in 2nd frame);
4. Samples from Our method are "conditionally diverse”. This is unseen in previous papers. E.g. when the egocentric camera sees only one arm, Ours will generate samples with this arm doing the motion seen (not perfectly accurate partially due to CLIP) and generate motions for the unseen arm and legs with diversity (see arms in 1st\&2nd frames on the 1st sequence, see legs in all frames on the second sequence).

\subsection{Variation of an error depending on the activity}
\label{suppsec:errvar}
\begin{figure*}[h]
  \centering
  \includegraphics[width=1\textwidth]{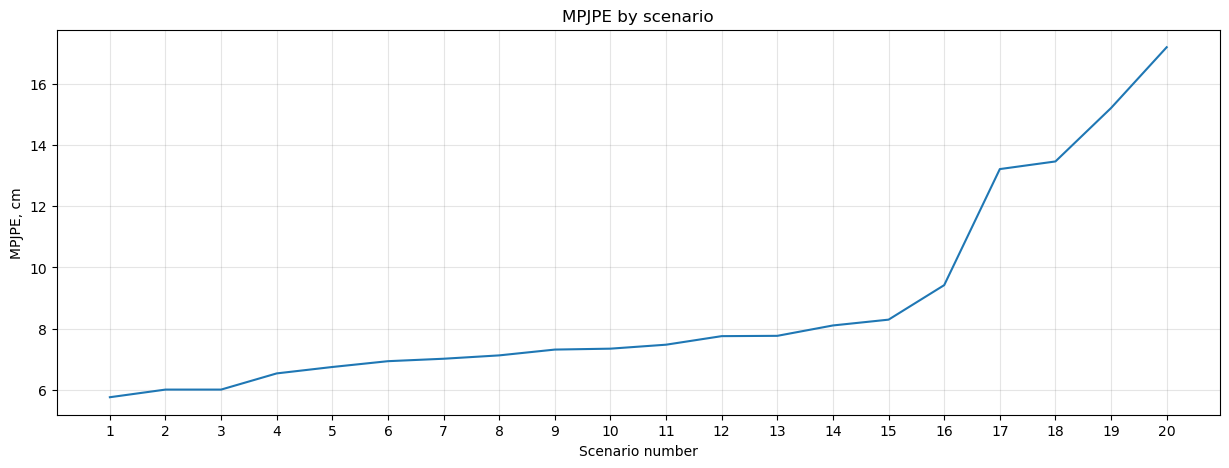}
\caption{MPJPE depending on the action scenario (sorted in increasing order).}
  \label{suppfig:mpjpe_by_script}
\end{figure*}

Our test dataset consists of diverse activities, and each sequence is dedicated to a certain type of activity according to the assigned scenario. In total, there are 20 scenarios, with indoor and outdoor activities featuring walking, sitting, laying, exercising, interacting with household objects, playing sports games, and more. If we group the sequences and measure the MPJPE in each group (Fig.~\ref{suppfig:mpjpe_by_script}), we can observe that the error is not distributed evenly -- while for most scenarios the error does not exceed 8cm, there is a chunk of challenging scenarios that have an error almost twice as high. To understand the reasons behind this, we selected and studied different metrics for the scenario, including the best, the worst, and median MPJPE. Results are presented in tables \ref{table:script_best}, \ref{table:script_med}, \ref{table:script_worst}. 

The best-performing scenario (Tab.~\ref{table:script_best}) consists of multi-terrain outdoor walking (hiking up and downhill) but does not feature any interactions. Small lower body error demonstrates that multi-level motion is, in general, not a significant challenge for our method -- in contrast to AvatarPoser, whose lower body error is higher on this scenario than on the mostly flat scenario from Tab.~\ref{table:script_med}. 

The scenario with the median method performance (Tab.~\ref{table:script_med}) consists of mostly flat-ground indoor multi-room interactions with the objects in the house (grabbing clothes, throwing pillows, opening doors). The subject often stays in the standing position, occasionally bending to reach some objects. As interactions with the objects appear more often here, we notice higher hand positional errors for our method. This can be explained by the inability of the CLIP-encoded image features to localize the hands precisely during the interactions. Occasional bending can also be misinterpreted for a different motion sometimes, which explains higher floor penetration error.

The worst performing scenario (Tab.~\ref{table:script_worst}) consists mainly of yoga and body stretching motions, which proved to be the most challenging for all the methods. While the upper body error is higher than usual, the error is primarily increasing due to very high lower body error. This is caused by a high position uncertainty: most of the time, lower body parts are not observed by the camera, and the floor estimation from a SLAM point cloud might be noisy. Future work on improving the performance in such scenarios might benefit from: enhancing the reconstructed SLAM pointcloud quality to provide reliable terrain information; including more of these challenging motions in the dataset; using cameras with a higher field of view, like fisheye cameras, to increase the body parts visibility.

\begin{table*}[ht]
\centering
\scriptsize
\renewcommand{\arraystretch}{1.3}
\begin{tabular}{l|c|c|c|c|c}
\toprule

& MPJPE $\downarrow$ & Hand PE $\downarrow$ & Low. PE $\downarrow$ & Up. PE $\downarrow$ & Floor Pen. $\downarrow$ \\ \hline

EgoEgo & \et{12.06}{0.33} & \et{31.31}{1.13} & \et{17.24}{0.75} & \et{9.40}{0.30} & \etb{0.01}{0.00} \\
AvatarPoser (Head) & 7.39 & 14.81 & 12.58 & 4.64 & 0.11 \\
Ours ($h=180$) & \etb{5.75}{0.03} & \etb{11.98}{0.13} & \etb{8.84}{0.07} & \etb{4.06}{0.03} & \et{0.02}{0.00} \\
Ours ($h=10$) & \et{6.19}{0.04} & \et{12.16}{0.07} & \et{9.97}{0.10} & \et{4.13}{0.01} & \et{0.02}{0.00} \\

\hline
\end{tabular}
\caption{Results for the scenario with the best HMD$^2$ performance. Scenario is consisting of the multi-terrain outdoor walking (hiking up- and downhill), mostly sightseeing. All the metrics are in cm.}
\label{table:script_best}

\centering
\scriptsize
\renewcommand{\arraystretch}{1.3}
\begin{tabular}{l|c|c|c|c|c}
\toprule

& MPJPE $\downarrow$ & Hand PE $\downarrow$ & Low. PE $\downarrow$ & Up. PE $\downarrow$ & Floor Pen. $\downarrow$ \\ \hline

EgoEgo & \et{12.29}{0.25} & \et{32.32}{0.50} & \et{16.40}{0.64} & \et{10.16}{0.16} & \etb{0.31}{0.14} \\
AvatarPoser (Head) & 8.39 & 20.94 & 11.44 & 6.78 & 0.80 \\
Ours ($h=180$) & \etb{6.53}{0.06} & \etb{15.66}{0.17} & \etb{8.86}{0.10} & \etb{5.29}{0.05} & \et{0.42}{0.05} \\
Ours ($h=10$) & \et{7.32}{0.05} & \et{17.30}{0.17} & \et{10.05}{0.10} & \et{5.87}{0.04} & \et{0.45}{0.02} \\

\hline
\end{tabular}
\caption{Results for the scenario with the median across all 20 scenarios HMD$^2$ performance. Scenario is consisting of flat-ground indoor multi-room interactions with the objects in the house (grabbing clothes, throwing pillows, opening doors), mostly upright standing with occasional bending (to reach for the next object). All the metrics are in cm.}
\label{table:script_med}

\centering
\scriptsize
\renewcommand{\arraystretch}{1.3}
\begin{tabular}{l|c|c|c|c|c}
\toprule

& MPJPE $\downarrow$ & Hand PE $\downarrow$ & Low. PE $\downarrow$ & Up. PE $\downarrow$ & Floor Pen. $\downarrow$ \\ \hline

EgoEgo & \et{28.67}{1.97} & \et{42.85}{1.46} & \et{52.11}{4.52} & \et{15.75}{0.64} & \et{12.76}{3.55} \\
AvatarPoser (Head) & 23.30 & 31.11 & 45.01 & 11.32 & 21.79 \\
Ours ($h=180$) & \etb{17.21}{0.20} & \etb{24.39}{0.36} & \etb{31.27}{0.50} & \etb{9.45}{0.13} & \etb{3.32}{0.24} \\
Ours ($h=10$) & \et{18.74}{0.65} & \et{26.28}{0.50} & \et{33.37}{1.41} & \et{10.55}{0.27} & \et{5.01}{0.39} \\

\hline
\end{tabular}
\caption{Results for the scenario with the worst HMD$^2$ performance. Scenario is consisting of challenging body stretching and yoga motions, mostly on done the floor, recorded indoors. All the metrics are in cm.}
\label{table:script_worst}
\end{table*}

\section{Limitations, future work and ethical implications}
\label{sec:supp_discussion}
As mentioned in the main paper, our system is limited by the data encoded in the features - the limbs localization precision is less than desired sometimes. Features that contain more precise positional information than CLIP may improve performance: one potential direction for future work is to additionally condition the method on the results of the hand-tracking algorithm. However, even without explicit positional information, CLIP-encoded images improve upper body tracking. The effect on the lower body is less apparent. This, of course, can be explained by the fact that the lower body is much less visible from the camera, especially since we use a camera with the standard FOV looking outwards. Additional information from the downward-looking wide-angle cameras can improve the performance, as shown in \eg~\cite{wang2023scene}. 

Even with the point cloud context provided, our method can sometimes produce visual artifacts such as floor penetration (as measured by the Floor. Pen. metric in tables). This means that the network occasionally misses or ignores the PC context. It can happen due to the noise presented in the pointcloud data and large distances between the points, especially in untextured regions like floors or walls. 
One way to improve the performance here is to use the more advanced point cloud/mesh reconstruction solution, potentially using the depth sensor (\eg Depth-based fusion~\cite{izadi2011kinectfusion}). Another way is to use a more advanced point cloud encoder; such an encoder can be trained on a different task, \eg, point-to-mesh~\cite{chibane2020neural}. Note that we only capture static point clouds and do not yet handle dynamic environment changes such as opening doors, moving a chair, etc. -- this is a great future work direction.

Our method is not aware of the shape of the body and, therefore, does not correct self-interpenetration of body parts, which can happen sometimes. That can be fixed during the postprocessing stage with self-contact optimization methods like TUCH~\cite{muller2021self}. Another problem that affects the visual quality is motion jitter, which can be observed mostly during online low-latency inference -- this can be smoothed during motion postprocessing. However, we decided not to apply the smoothing to show the raw performance of the method.

As our method uses the head-mounted first-person view camera, there are privacy concerns related to that; one of the major ones is the leaking of the raw video frames. Our current effort to mitigate this involves using the built-in functionality of Aria glasses~\cite{aria23surreal} to blur the faces during the data capture. We can improve the privacy aspect even more by moving CLIP and PC encoding computation on the capturing device itself. As our method uses only the encoded image and pointcloud features instead of raw data, on-device precomputed features would work just as well. We also believe that after some optimization efforts, there is a potential to perform the full inference pipeline on the mobile device itself, therefore eliminating the potential data leak problem completely.

\bibliographystyle{splncs04}
\bibliography{main}
\end{document}